\documentclass[11pt,a4paper]{article}
\usepackage[hyperref]{acl2020}
\usepackage{times}
\usepackage{latexsym}
\usepackage{graphicx}
\usepackage{xspace}
\usepackage{booktabs}
\usepackage{graphicx}
\usepackage{multirow}
\usepackage{xcolor}
\usepackage{subcaption}
\usepackage{amsmath}

\aclfinalcopy
\def\aclpaperid{546}

\title{Closing the Gap: Joint De-Identification and Concept Extraction \\in the Clinical Domain}

\author{Lukas Lange$^{1,2,3}$ \\
	\And
	Heike Adel$^1$ \\
	$^1$ Bosch Center for Artificial Intelligence, Renningen, Germany\\
	$^2$ Spoken Language Systems (LSV), Saarland University, Saarbr\"{u}cken, Germany\\
	$^3$ Saarbr\"{u}cken Graduate School of Computer Science, Saarbr\"{u}cken, Germany\\
	{\tt \{Lukas.Lange,Heike.Adel,Jannik.Stroetgen\}@de.bosch.com} \\
	\And
	Jannik Str\"{o}tgen$^1$ \\
	\\}

\date{}

\begin{document}
	\maketitle
	
	\begin{abstract}
Exploiting natural language processing in the clinical domain requires de-identification, i.e., anonymization of personal information in texts.
However, current research considers de-identification and downstream tasks, such as concept extraction, only in isolation and does not study the effects of de-identification on other tasks.
In this paper, we close this gap
by reporting concept extraction performance on automatically anonymized data and investigating joint models for de-identification and concept extraction.
In particular, we propose a stacked model with restricted access to privacy-sensitive information and a multitask model.
We set the new state of the art on benchmark datasets in English (96.1\% F1 for de-identification and 88.9\% F1 for concept extraction) and Spanish (91.4\% F1 for concept extraction).
\end{abstract}	
	\section{Introduction}
\label{sec:introduction}
In the clinical or biomedical domain, natural language processing (NLP) could significantly improve the efficiency and effectiveness of processes. 
For example, the extraction of structured information from clinical narratives can help in decision making or drug repurposing \cite{meddocan/task/marimon2019}.
However, the automatic processing of documents with privacy-sensitive content is restricted due to the necessity of applying anonymization techniques.

\begin{figure}
	\centering
	\includegraphics[width=.43\textwidth]{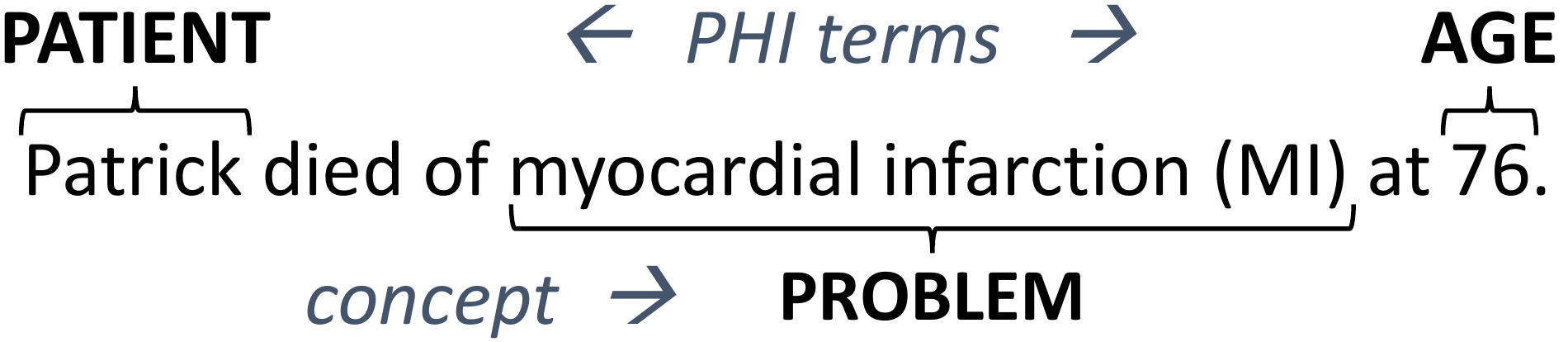}
	\caption{Sentence with annotations of the two tasks.}
	\label{fig:examples}
\end{figure}

Text anonymization, also called de-identification, aims at detecting and replacing protected health information (PHI),\footnote{PHI types are typically defined by governments, for instance in the Health Insurance Portability and Accountability Act (HIPAA) of the United States.} such as patient names, age and phone numbers, as shown in the upper part of Figure \ref{fig:examples}.
Recent studies show that automatic de-identification leads to promising results \cite{i2b2/task/stubs2015,meddocan/task/marimon2019}.
A severe limitation of current approaches in research, however, is that de-identification is typically addressed in isolation but not together with a downstream task, such as concept extraction (CE) from medical texts \cite{pharmaconer/task/gonzalez2019, i2b2/task/uzuner2010}.
Instead, the downstream task models are trained and evaluated on the non-anonymized data, and it remains unclear how de-identification affects their performance in real-world settings. 

In this paper, we argue that to evaluate the effectiveness of NLP in the medical domain, the tasks of de-identification and information extraction should be analyzed together.
Our contributions are as follows:
We close this gap and analyze the effect of de-identification on clinical concept extraction. 
Moreover, we consider the two tasks jointly and propose two end-to-end models: 
A multitask model that shares the input representation across tasks, and a stacked model that trains a pipeline of de-identification and concept extraction in an end-to-end manner.
For the stacked model, we propose to use a masked embedding layer to restrict the access of the concept detector to privacy-sensitive information and train it on an anonymized version of the data.
To make the model differentiable, we use the Gumbel softmax trick \cite{gumbel/maddison2016,gumbel/jang2016}.

	\begin{figure*}[t]
	\centering
	\begin{subfigure}[t]{0.303\textwidth}
		\centering
		\includegraphics[width=1.0\textwidth]{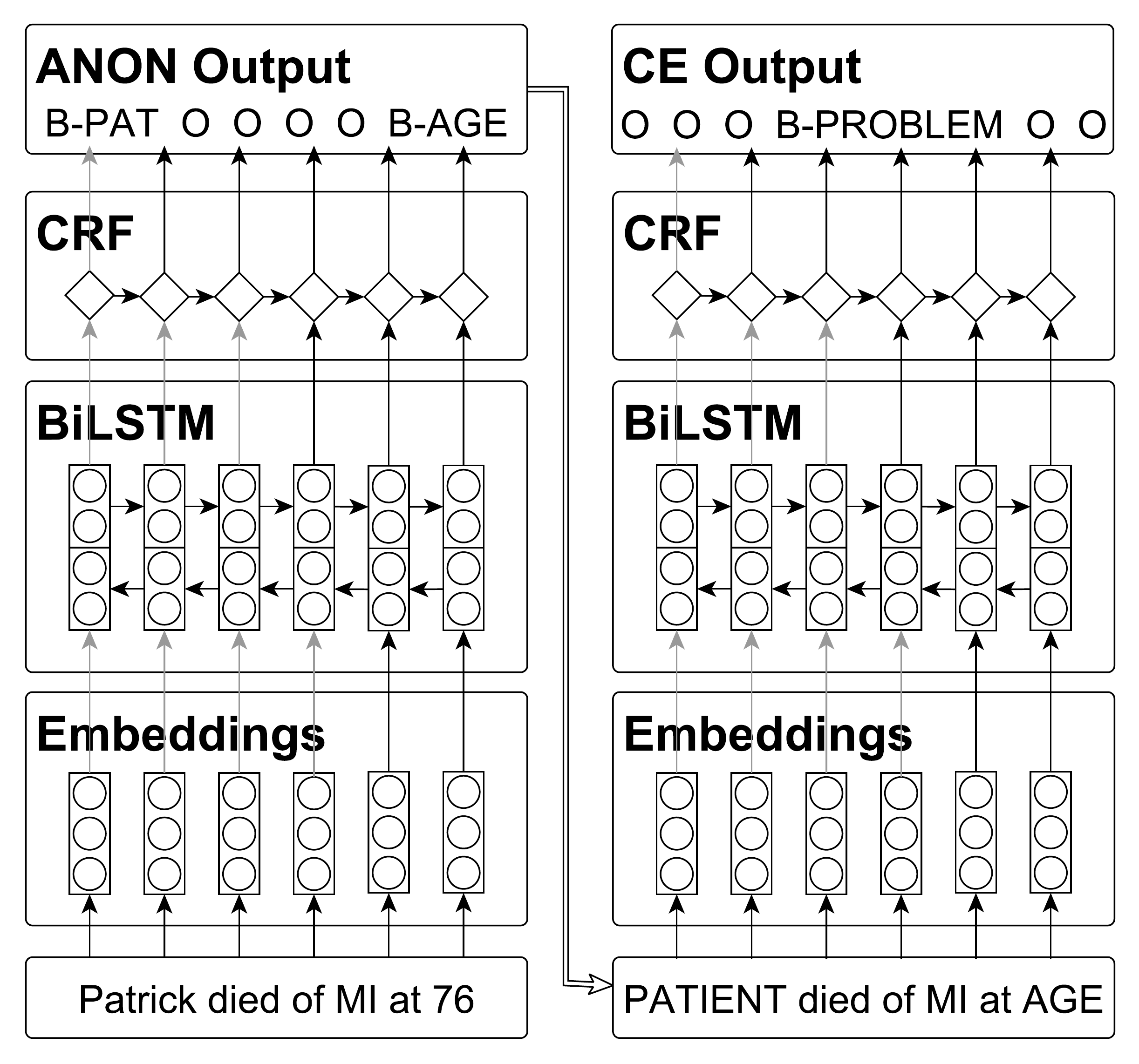}
		\caption{Pipeline Model.}
		\label{fig:model_pipeline}
	\end{subfigure}
	\enskip
	\begin{subfigure}[t]{0.291\textwidth}
		\centering
		\includegraphics[width=1.0\textwidth]{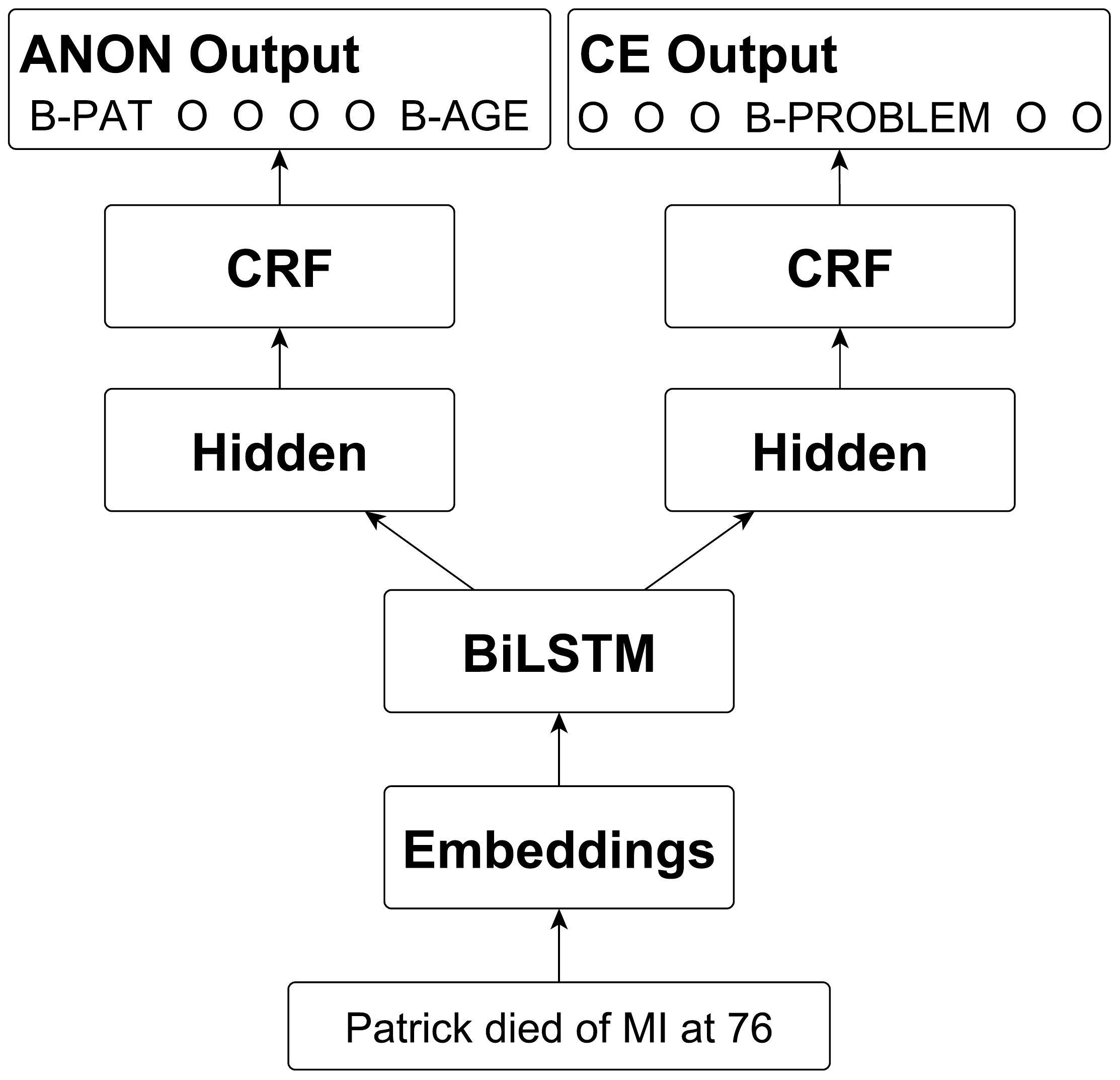}
		\caption{Multitask Model.}
		\label{fig:model_multitask}
	\end{subfigure}
	\enskip
	\begin{subfigure}[t]{0.365\textwidth}
		\centering
		\includegraphics[width=1.0\textwidth]{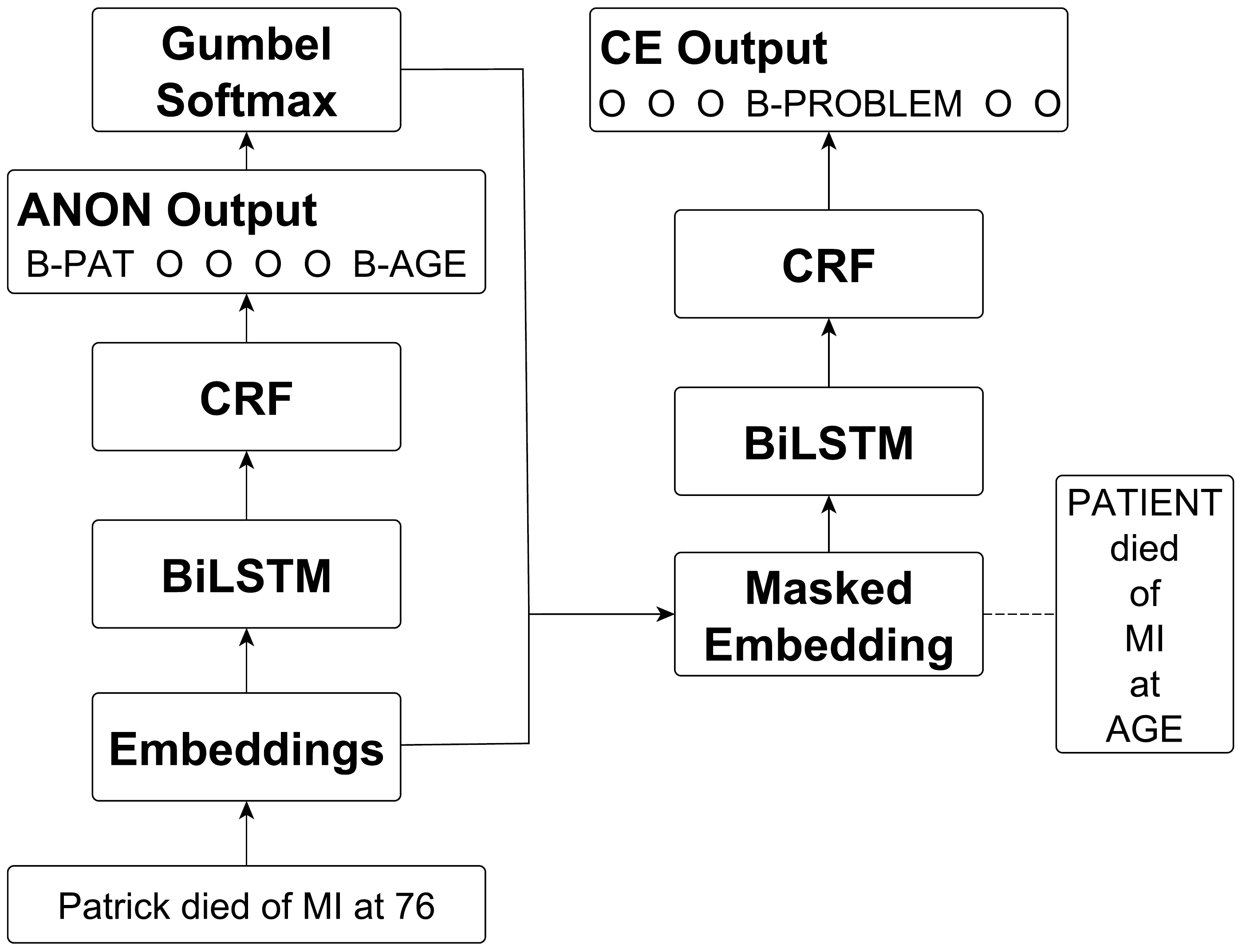}
		\caption{Stacked Model.}
		\label{fig:model_joint}
	\end{subfigure}
	\caption{Overview of our different model architectures.
		While the CE model in the multitask setting has access to all privacy-sensitive information, the access of the pipeline CE model and the stacked CE model is restricted by the ANON output.
		``PAT'' stands for Patient. The labels are encoded in BIO format.}
	\label{fig:models}
\end{figure*}

We conduct experiments on clinical benchmark datasets in English and Spanish. 
Our results indicate that de-identification does not affect CE models negatively, but has even a slight positive effect on the results, probably because de-identification homogenizes the input for CE.
Modeling both tasks jointly leads to better results than treating de-identification as a pure preprocessing step, resulting in new state-of-the-art performance for CE.

For future research, we publish our code.\footnote{\url{https://github.com/boschresearch/joint_anonymization_extraction}}

	\section{Related Work}
\label{sec:relatedWork}
While many works propose joint training for other NLP tasks \cite[i.a.,][]{joint/finkel-manning-2009,miwa-sasaki-2014-modeling},
including multitask learning \cite[i.a.,][]{collobert2008unified,klerke-etal-2016-improving,sogaard-goldberg-2016-deep} and stacking of pipeline components \cite[i.a.,][]{miwa-bansal-2016-end}, we are to the best of our knowledge the first to combine de-identification with an information extraction task. 
In this section, we report related work in those two fields.

\subsection{De-Identification}
The increasing importance of de-identification is reflected in the number of shared tasks \cite{i2b2/task/uzuner2007,i2b2/task/stubs2015,meddocan/task/marimon2019}.
State-of-the-art methods for de-identification  typically rely on recurrent neural networks (RNNs) \cite{i2b2/dernoncourt2016,meddocan/lange2019,kajiyama-etal-2018-de}.
	
\newcite{anon/feutry2018} and \newcite{i2b2/friedrich2019} 
create pseudo-de-identified text representations with adversarial training. In particular, they replace personal information, such as names, by other names.
\newcite{i2b2/zhao2018} augment the training data by creating more general text skeletons, e.g., by replacing rare words, such as names, by a special unknown token.
Compared to these works, we
exploit the advantages of both approaches
and replace personal information by their class names as placeholders.
This approach is not only common for de-identification~\cite{johnson2016mimic}, but also for relation extraction where entities are often either replaced by their type or enriched with type information \cite[i.a.,][]{zhang-etal-2017-position,miwa-sasaki-2014-modeling}. We further motivate our choice in Section~\ref{sec:pipeline}.
Another difference to the above mentioned works is that we do not augment the training data for our de-identification model.

\subsection{Medical Information Extraction}
Analogously, there have been a series of shared tasks for information extraction in the clinical and biomedical domain  \cite{i2b2/task/uzuner2010,i2b2/task/sun2012,biocreative/task/krallinger2015,pharmaconer/task/gonzalez2019}.
Models for these tasks often either rely on hand-crafted features \cite{biocreative/leaman2015, i2b2/2010/xu2012} or RNNs \cite{biocreative/hemati2019, biocreative/korvigo2018, tourille-etal-2018-evaluation}.
\citet{newman-griffis-zirikly-2018-embedding} study the performance of RNNs for medical named entity recognition in the context of patient mobility and find that they benefit from domain adaption.
	
In contrast to previous work, we investigate the usage of de-identified texts as input for clinical concept extraction models and propose to jointly model de-identification and concept extraction.
	\section{Model}
\label{sec:model}
In this section, we present our systems for the two individual tasks and our proposed joint models. 
Figure~\ref{fig:models} shows the respective architectures.

\subsection{General Architecture}
\label{sec:classification}
We model both document anonymization (ANON) and clinical concept extraction (CE) as sequence labeling problems and apply 
a bidirectional long short-term memory (BiLSTM) network~\cite{model/hochreiter1997} with a conditional random field (CRF) output layer~\cite{crf/lafferty2001}, similar to \newcite{model/lample2016}.
In recent works on clinical de-identification and CE, this architecture was shown to be very promising \cite{meddocan/task/marimon2019, pharmaconer/task/gonzalez2019}.

Each token is represented with a concatenation of different pre-trained language-specific embeddings:
byte-pair-encoding \cite{emb/heinzerling2018}, fastText \cite{emb/bojanowski2017} and FLAIR \cite{emb/akbik2018}. For Spanish, we also include multilingual BERT embeddings~\cite{emb/devlin2018}. Further, we include the following domain-specific embeddings: clinical BERT for English pre-trained on discharge summaries \cite{i2b2/alsentzer2019} and clinical fastText for Spanish pre-trained on the Spanish E-health corpus from the Scielo archive \cite{emb/soares2019}.

\subsection{Pipeline Model}
\label{sec:pipeline}
To assess the effects of de-identification on CE, we first apply the de-identification model to anonymize the CE dataset and then evaluate the CE model on the anonymized data. 
We refer to this approach as \textsc{Pipeline} model (see Figure~\ref{fig:model_pipeline}). 
For anonymization, we replace each detected privacy-sensitive term with a placeholder of its PHI type, i.e., there is one placeholder per type.

This replacement choice has advantages over the alternatives described in Section \ref{sec:relatedWork}.
Compared to replacing personal information with alternative names,
it leads to a more general text and thus, homogenizes the input for the downstream-task classifier.
Compared to replacing all personal information with the same token, the resulting text is more specific, allowing the downstream-task classifier to take into account which kind of personal information was mentioned.
Thus, the approach is a trade-off between more homogeneous input and more fine-grained information for the downstream-task classifier.

\subsection{Joint Models}
\label{sec:joint}
Instead of using a sequential pipeline, we propose to train both tasks jointly. For this, we test two approaches: 
a multitask model and a stacked model.

\subsubsection{Multitask Model}
In the \textsc{Multitask} model (Figure~\ref{fig:model_multitask}), the weights up to the BiLSTM layer are shared across both tasks. For each task, we add a task-specific hidden layer with ReLU activation and a CRF output layer.
Note that in this architecture, the CE model has access to the original, privacy-sensitive data.

\subsubsection{Stacked Model}
We also propose a \textsc{Stacked} model (Figure~\ref{fig:model_joint}), where only the de-identification part has access to the privacy-sensitive information. The access of the CE part is restricted by a masked embedding layer as described in the following.

\begin{figure}
	\centering
	\includegraphics[width=.48\textwidth]{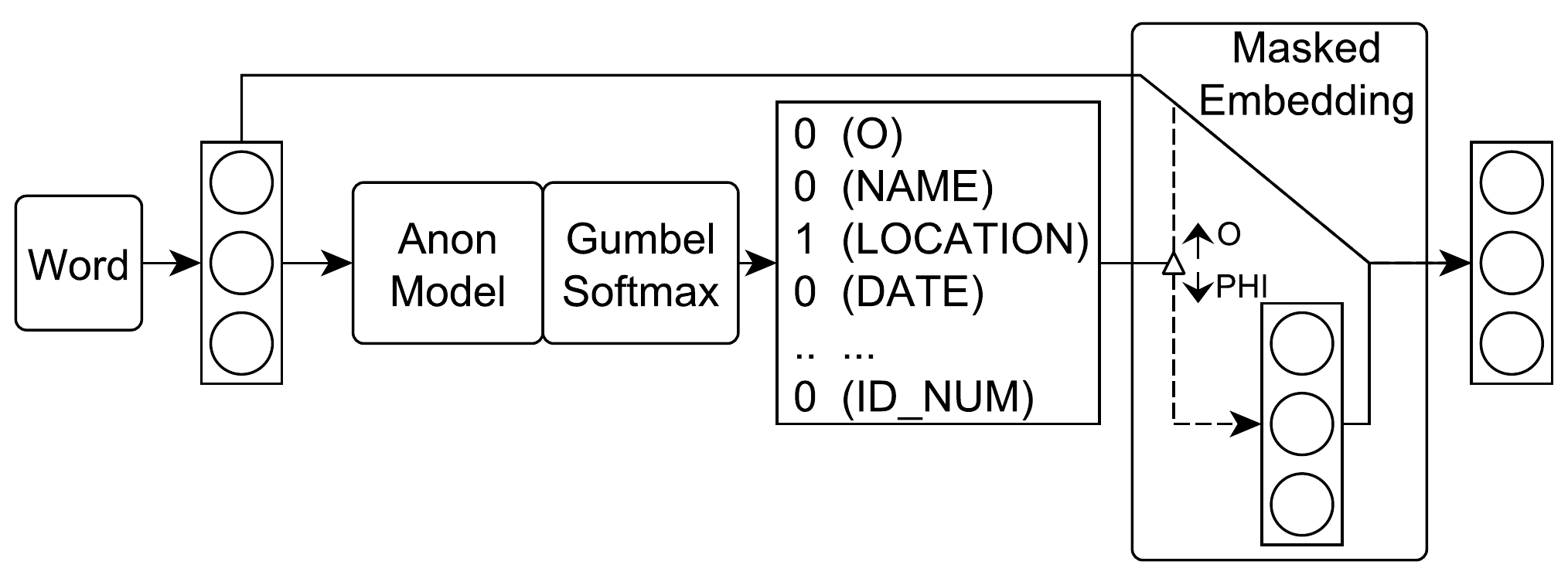}
	\caption{Masked Embedding.}
	\label{fig:masked_embedding}
\end{figure}

\textbf{Masked Embedding Layer.}
The masked embedding layer ensures that the CE model does not have access to privacy-sensitive information by replacing the input embeddings of privacy-sensitive tokens by PHI-class embeddings which are randomly initialized and fine-tuned during training. 
This is depicted in Figure \ref{fig:masked_embedding}.

\textbf{Gumbel Softmax Trick.}
The masked embedding layer requires a discrete output from the de-identification part.
In order to ensure that the model stays fully differentiable, we use the Gumbel softmax trick \cite{gumbel/maddison2016, gumbel/jang2016}.
It approximates categorical samples with a continuous distribution on the simplex and computes gradients for backpropagation with the reparameterization trick.
The Gumbel softmax function has the following form:

\vspace*{-0.25ex}

\begin{equation}
\begin{aligned}
y_k^\tau =
\frac{\exp((\log \alpha_k +G_k)/\tau)}{\sum_{i=1}^K \exp((\log \alpha_{i} +G_{i})/\tau)}
\end{aligned}
\end{equation}
with $\alpha_1,...\alpha_K$ being the unnormalized output scores from the de-identification layer, $G_1,...,G_K$ being i.i.d samples drawn from Gumbel(0, 1) and $\tau$ being a temperature. 
For $\tau \rightarrow 0$, the distribution becomes identical to the categorical distribution.

The masked embedding layer takes the output of the Gumbel softmax (i.e., an anonymization label) and if the label is a PHI class and requires anonymization, the masked embedding layer uses the respective PHI class embedding vector, otherwise it uses the original embedding vector.

	\section{Experiments}
\label{sec:experiments}
In this section, we describe the datasets used in our experiments, and training details for our models. Finally, we present our results and analysis.

\begin{table}
	\footnotesize
	\centering
	\begin{tabular}{l|rr|rr} \toprule
		& \multicolumn{2}{c|}{English (i2b2)} & \multicolumn{2}{c}{Spanish} \\
		& ANON & CE & ANON & CE\\
		\midrule
		\# classes & 24 & 3 & 22 & 3\\
		train (\# tokens) & 45,793 & 16,315 & 15,903 & 8,068\\
		dev (\# tokens) & 5,088 & - & 8,277 & 3,748\\
		test (\# tokens) & 32,587 & 27,625 & 7,966 & 3,930\\
		\bottomrule
	\end{tabular}
	\caption{Dataset statistics. \# classes denotes the number of classes without the negative `O' class. }
	\label{tab:datasets}
\end{table}

\subsection{Data and Model Training and Evaluation}
\label{sec:data}
We evaluate our models on corpora from the clinical domain in English and Spanish.
For English, we use the data from the i2b2 2010 CE task \cite{i2b2/task/uzuner2010} and the i2b2 2014 de-identification task \cite{i2b2/task/stubs2015}.
For Spanish, we use the MEDDOCAN \cite{meddocan/task/marimon2019} corpus for de-identification and the PharmaCoNER corpus~\cite{pharmaconer/task/gonzalez2019} for CE. 
As PharmaCoNER is a subset of MEDDOCAN, we have both gold-standard concept and de-identification annotations for this data.

\textbf{Data Preprocessing. }
We used the preprocessing scripts from~\newcite{i2b2/alsentzer2019} for the English i2b2 corpora and the Spanish Clinical Case Corpus tokenizer~\cite{spaccc/Intxaurrondo19} for both Spanish corpora. We noticed that the Spanish tokenizer sometimes merges multi-word expressions into a single token joined with underscores for contiguous words. As a result, some tokens cannot	be aligned with the corresponding entity annotations. To address this, we split those tokens into their components in a postprocessing step.
Table~\ref{tab:datasets} shows statistics about corpora sizes.

\textbf{Hyperparameters. }
The embeddings have 300 (byte-pair-encoding), 300 (fastText)
and 4,048 (FLAIR) dimensions. 
For English, we use clinical BERT embeddings with 768 dimensions which are constructed by averaging the last four layers with the scalar mix operation proposed by~\newcite{liu-etal-2019-linguistic}.
We concatenate all embeddings to one input vector, resulting in a total input dimensionality of 5,416. 
Analogously, we use multilingual BERT (768 dim.) and domain-specific fastText embeddings (100 dim.) for Spanish, resulting in 5,516 input dimensions. 
For the LSTM, we use 256 hidden units per direction.
The task-specific hidden layer of the multitask model has 128 units.

\textbf{Training.}
For training, we use stochastic gradient descent with a learning rate of 0.1 and a batch size of 32 sentences. The learning rate is halved after 3 consecutive epochs without improvements on the development set.
For the joint models, we pretrain the anonymization part for 3 epochs and, then, use a higher initial learning rate of 0.2 for the concept extraction part. 
We perform early stopping on the development set. If no development set was provided by the corpus (i2b2 2010 corpus), we held out 10\% of the training set as development set.
Note that we use the same hyperparameters for all our models and all tasks, which were tuned on the Spanish concept extraction data. 

\textbf{Evaluation. }
We train each model with three random initializations and report F1 for exact matching for the best model in all experiments.
We perform statistical significance testing to check if our joint models are better than the \textsc{Pipeline} model. 
We use paired permutation testing with 2$^{20}$ permutations and a significance level of 0.05. 

\begin{table}
	\footnotesize
	\centering
	\begin{tabular}{l|c|c}\toprule
		Models & English & Spanish \\ \midrule
		\newcite{i2b2/yang2015}       & 96.0 & \\
		\newcite{i2b2/zhao2018}       & 94.0 & \\
		\newcite{i2b2/alsentzer2019}  & 93.0 & \\
		\newcite{meddocan/lange2019}  & & \bf 97.0 \\
		\newcite{meddocan/hassan2019} & & 96.3 \\
		\newcite{meddocan/perez2019}  & & 96.0 \\
		\midrule
		Our \textsc{Pipeline} (ANON only)      & \bf 96.1 & 96.8 \\ 
		Our \textsc{Stacked}          & 95.9 & 96.8 \\
		Our \textsc{Multitask}        & 95.2$^\ast$ & 96.7 \\ 
		\bottomrule
	\end{tabular}
	\caption{F1 results for de-identification.
		$^\ast$ highlights our models with statistically significant differences compared to \textsc{Pipeline} (ANON only). }
	\label{tab:anon}
\end{table}

\subsection{Results}
Table~\ref{tab:anon} shows that the de-identification component of our \textsc{Pipeline} model which was trained on the single task of de-identification
sets the new state of the art on English and performs comparable on Spanish.
The performance difference to our prior work \cite{meddocan/lange2019} is due to a slightly different set of input embeddings. However, we found no statistically significant differences to that model. 
The de-identification performance
of \textsc{Stacked} is comparable to the \textsc{Pipeline} model, the de-identification performance of \textsc{Multitask} is slightly lower, however, we only found statistically significant differences for the \textsc{Multitask} model for English. 

The results for our concept extraction models in comparison to state of the art are shown in Table~\ref{tab:ner}.
We set the new state of the art on both languages.
While in the \textsc{Pipeline} setting, the CE performance is slightly lower 
(as it has been trained on the non-anonymized texts but is evaluated on the de-identification output), training de-identification and CE jointly leads to considerable improvements for both \textsc{Stacked} and \textsc{MultiTask}
with statistically significant differences for both models in English and for \textsc{Multitask} also in Spanish.
Especially the results of \textsc{Stacked} in comparison to \textsc{Pipeline} shows that end-to-end training of the two steps is promising, 
while still preserving privacy aspects during model training by restricting internal access to PHI tokens.
The performance of each embedding used in our experiments is shown in Table~\ref{tab:embed}. 
As mentioned before, we did not include multilingual BERT embeddings for English, but show their results for completeness. 

\begin{table}
    \footnotesize
    \centering
	\begin{tabular}{l|c|c|c} \toprule
		Models & No PHI & English & Spanish \\  \midrule
		\newcite{i2b2/2010/bruijn2010}   & no & 85.2 & \\ 
		\newcite{i2b2/2010/xu2012}       & no & 84.9 & \\ 
		\newcite{i2b2/alsentzer2019}     & no & 87.7 & \\
		\newcite{pharmaconer/sun2019}    & no & & 89.2 \\
		\newcite{pharmaconer/lange2019}  & no & & 88.6 \\
		\midrule
		Our \textsc{Pipeline}   & \textbf{yes} & 88.0 & 89.6 \\
		Our \textsc{Stacked}    & \textbf{yes} & 88.7$^\ast$ & 90.0 \\
		Our \textsc{Multitask}  & no & \bf 88.9$^\ast$ & \bf 90.3$^\ast$ \\
		\midrule
		\newcite{pharmaconer/xiong2019}$^\dagger$    & no & & 91.1 \\
		\newcite{pharmaconer/stoeckel2019}$^\dagger$ & no & & 90.5 \\
		Our \textsc{Multitask}$^\dagger$             & no & & \bf 91.4$^\ast$ \\ 
		\bottomrule
	\end{tabular}
	\caption{F1 results for concept extraction. 
	We indicate for each model whether anonymized data is used during the extraction training. 	
	$^\dagger$ indicates models which are trained on a combination of training and development set.
	$^\ast$ highlights our models with statistically significant differences compared to our \textsc{Pipeline}. }
	\label{tab:ner}
\end{table}

\begin{table}
	\footnotesize
	\centering
	\begin{tabular}{l|c|c} \toprule
		Embedding & English & Spanish \\  \midrule
		fastText     & 81.5 & 78.7 \\
		byte-pair-encoding        & 83.4 & 83.9 \\
		FLAIR        & 83.0 & 82.4 \\
		Multilingual BERT & 84.4 & 85.9 \\
		Clinical BERT (English) & 87.2 & - \\
		Clinical fastText (Spanish) & - & 79.7 \\
		Concatenation of all   & 88.1 & 89.7 \\
		\bottomrule
	\end{tabular}
	\caption{Effects of different embeddings on the concept extraction task (without anonymization).  }
	\label{tab:embed}
\end{table}

\subsection{Analysis of Pipeline Setting}
\begin{table}
	\footnotesize
	\centering
	\begin{tabular}{l|l|l|l|l} \toprule
	 &	train on & test on & dev & test\\
	\midrule
	(1) &	non-anon. & non-anon. & 89.2 & 89.7 \\
	(2) &	non-anon. & anon-predicted & 89.1 & 89.6 \\	
	(3) &	anon-predicted & non-anon. & 89.2 & 89.6 \\
	(4) &	anon-predicted & anon-predicted & 89.6 & 90.0 \\
	(5) & 	anon-gold & anon-predicted & 89.5 & 90.0 \\
	(6) &	anon-gold & anon-gold & 89.6 & 90.1 \\ 
	\bottomrule
	\end{tabular}
	\caption{Pipeline analysis results on Spanish concept extraction. ``non-anon.'' indicates the original text without anonymization; ``anon-gold'' and ``anon-predicted'' refer to texts with replacements for gold/predicted de-identification labels, respectively.}
	\label{tab:analysis}
\end{table}

Finally, we analyze the impact of de-identification on CE. 
The results for training and testing our CE model on different inputs (non-anonymized vs.\ anonymized) are shown in Table \ref{tab:analysis}.
We restrict our analysis to Spanish since the data is labeled with both de-identification and concept information (see Section \ref{sec:data}).
Thus, we can also investigate the difference between gold and predicted de-identification labels.
The CE model benefits from being trained and evaluated on anonymized data (lines \mbox{4-6)}.
However, it hurts to train on non-anonymized data and evaluate on predicted de-identification labels (line 1 vs.\ 2) and vice versa (line 1 vs.\ 3). 
This supports our motivation that it is necessary to investigate anonymization and downstream applications together.
The difference of training on gold vs.\ predicted de-identification labels (lines \mbox{4-6)} is only marginal, suggesting that state-of-the-art de-identification systems are good enough to be used in such settings.

	\section{Conclusion}
In this paper, we close the gap and consider de-identification of clinical text together with concept extraction, a possible downstream application. We investigate the effects of de-identification on concept extraction and show that it positively influences the concept extraction performance. We propose two models to learn both tasks jointly, a multitask model and a stacked model, and set the new state of the art on medical concept extraction benchmark datasets for English and Spanish.

	\section*{Acknowledgments}
	We would like to thank the members of the BCAI NLP\&KRR research group and the anonymous reviewers for their helpful comments.
	
	\bibliographystyle{acl_natbib}	
	\bibliography{references}
	
\end{document}